%% file: root.tex
\documentclass[letterpaper, 10 pt, conference]{ieeeconf}  

\IEEEoverridecommandlockouts                              

\usepackage{amsmath,amsfonts,amssymb}
\usepackage{array}
\usepackage{textcomp}
\usepackage{stfloats}
\usepackage{url}
\usepackage{verbatim}
\usepackage{graphicx}
\usepackage{graphics} 
\usepackage{epsfig} 
\usepackage{times} 
\usepackage{amsmath} 
\usepackage{amssymb}  
\usepackage{booktabs}

\usepackage{pifont}
\usepackage{multirow}
\usepackage[table,xcdraw]{xcolor}

\usepackage[mathscr]{eucal}

\newcommand{\mytabref}[1]{Table~\ref{#1}}

\newcommand{\myfigref}[1]{Fig.~\ref{#1}}

\newcommand{\mysecref}[1]{Sec.~\ref{#1}}

\newcommand{\colorfirst}{\cellcolor[HTML]{c0e2ca}}
\newcommand{\colorsecond}{\cellcolor[HTML]{fff5b3}}
\newcommand{\colorthird}{\cellcolor[HTML]{ffd9b3}}

\def\method{Semi-SMD}

\title{\LARGE \bf
\method: Semi-Supervised Metric Depth Estimation via Surrounding Cameras for Autonomous Driving
}

\author{Yusen Xie, Zhengmin Huang, Shaojie Shen, and Jun Ma%
\thanks{Yusen Xie, Zhenmin Huang, and Jun Ma are with the Robotics and Autonomous Systems Thrust, The Hong Kong University of Science and Technology (Guangzhou), Guangzhou 511453, China {\tt\footnotesize yxie827@connect.hkust-gz.edu.cn; zhuangdf@connect.ust.hk; jun.ma@ust.hk}}
\thanks{Shaojie Shen is with the Department of Electronic and Computer Engineering, The Hong Kong University of Science and Technology, Hong Kong SAR, China {\tt\footnotesize eeshaojie@ust.hk}}}

\begin{document}

\maketitle
\thispagestyle{empty}
\pagestyle{empty}

\input{sec/0_abstract}

\input{sec/1_intro}
\input{sec/2_relatedwork}
\input{sec/3_method}
\input{sec/4_experiments}
\input{sec/5_conclusion}

\bibliographystyle{IEEEtran}
\bibliography{main}

\end{document}

%% file: sec/0_abstract.tex
\begin{abstract}
In this paper, we introduce \method, a novel metric depth estimation framework tailored for surrounding cameras equipment in autonomous driving.
In this work, the input data consists of adjacent surrounding frames and camera parameters. 
We propose a unified spatial-temporal-semantic fusion module to construct the visual fused features. 
Cross-attention components for surrounding cameras and adjacent frames are utilized to focus on metric scale information refinement and temporal feature matching.
Building on this, we propose a pose estimation framework using surrounding cameras, their corresponding estimated depths, and extrinsic parameters, which effectively address the scale ambiguity in multi-camera setups.
Moreover, semantic world model and monocular depth estimation world model are integrated to supervise the depth estimation, which improve the quality of depth estimation.
We evaluate our algorithm on DDAD and nuScenes datasets, and the results demonstrate that our method achieves state-of-the-art performance in terms of surrounding camera based depth estimation quality. 
The source code is available on GitHub{\footnote{https://github.com/xieyuser/Semi-SMD}}.
\end{abstract}

%% file: sec/1_intro.tex
\vspace{-0.5em}
\section{Introduction}
\label{sec:intro}
\vspace{-0.5em}
Metric depth estimation provides absolute distance perception of the surrounding environment in autonomous driving scenarios, supporting the following tasks such as motion forecasting~\cite{shi2025motion} and path planning~\cite{hu2023planning,teng2023motion}. Some existing framworks~\cite{geiger2012we,karnchanachari2024towards} directly detect depth by introducing LiDAR and Radar sensors, but these approaches lead to increased costs and algorithmic complexity. 
Currently, autonomous driving perception systems tend to favor visual-only surrounding camera solutions~\cite{hu2023planning,zhang2024sparsead} that are easier to implement. 
Some visual-based depth prediction algorithms~\cite{bhat2023zoedepth, yang2024depth,yang2024v2depth,godard2019digging,godard2017unsupervised} use large datasets to train monocular depth estimation (MDE), achieving impressive results in depth estimation as world models. 
However, the scale ambiguity encountered in monocular depth estimation makes it unsuitable for applications requiring precise depth information in autonomous driving. 
Surrounding cameras, which inherently provide scale information from extrinsic parameters, have seen rapid development in metric depth prediction~\cite{schmied2023r3d3,wimbauer2021monorec,bhat2023zoedepth}. 
These methods typically use data from the same frame~\cite{bhat2023zoedepth} or from the same camera~\cite{schmied2023r3d3,wimbauer2021monorec}, but do not take advantage of the stereo geometric constraints provided by spatial and temporal information, which results in insufficient accuracy and poor generalization.
Additionally, some methods~\cite{zou2024m,wei2023surrounddepth} use semi-supervised training guidance but face challenges in decoupling complex tasks efficiently. Therefore, the results are not accurate and easy to convergence to local optima.
Lastly, these methods fail to integrate semantic information effectively, leading to unclear boundaries in depth maps, particularly in areas where humans rely on semantic cues.

\begin{figure*}[!htb]
\centering
\includegraphics[width=\linewidth]{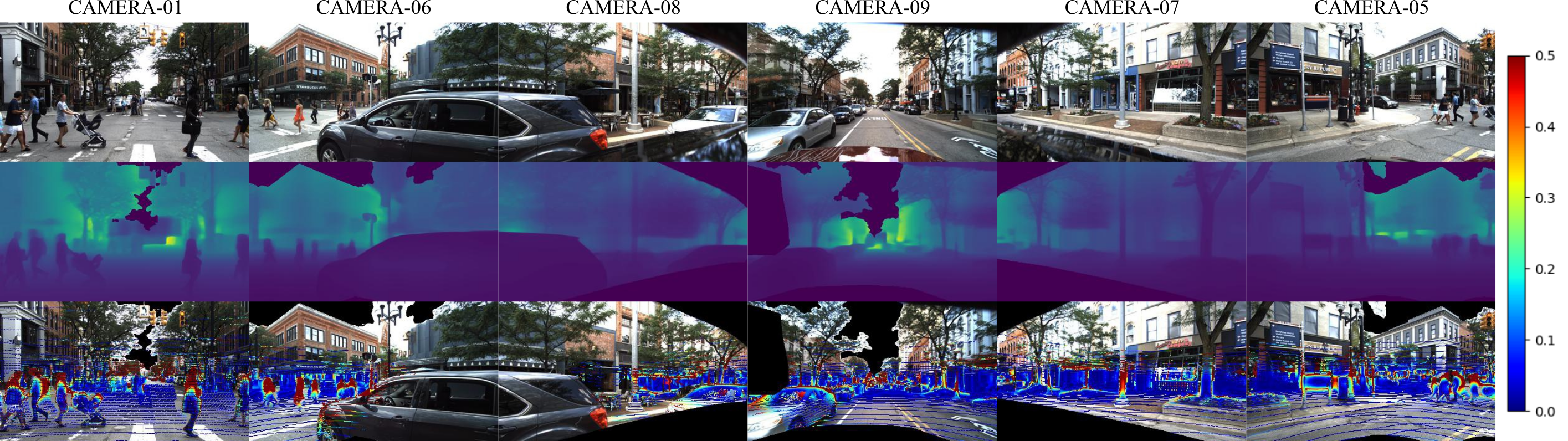}
\caption{We use adjacent surrounding camera images and the camera's intrinsic and extrinsic parameters as input, while simultaneously estimating the pose transformation between the two frames and the metric depth for each image. An illustration of our metric depth estimation results on the DDAD~\cite{guizilini20203d} dataset is shown. The first row displays the original surrounding RGB images, the second row shows the estimated metric depth, and the third row provides a quantitative visualization of \textit{Abs.Rel.} with the projected LiDAR points. The colors represent the error distribution.}
\label{fig:firstdemo}
\vspace{-1.5em}
\end{figure*}

To address these issues, we propose a \underline{\textbf{s}}urrounding camera \underline{\textbf{m}}etric \underline{\textbf{d}}epth estimation framework, named \textbf{\method}. The results of the paper are briefly presented in~\myfigref{fig:firstdemo}. By utilizing two adjacent frames of surrounding camera images, we employ a unified spatial-temporal-semantic transformer~\cite{vaswani2017attention} to fuse visual features extracted by ResNet~\cite{he2016deep} and semantic features from a semantic segmentation world model~\cite{zhang2023faster}. This fused features are then used for both depth prediction and pose estimation to boost computational efficiency.
Furthermore, we integrate depth prediction and the extrinsic parameters of the surrounding camera into the pose estimation network, redesigning a surrounding camera pose estimation module with precise scale information. Additionally, we introduce a curvature loss based on the depth estimation world model~\cite{yang2024depth, yang2024v2depth}. Experimental results demonstrate that the inclusion of this loss function significantly enhances the model’s convergence speed and depth prediction accuracy.
Our contributions are summarized as follows:

\begin{itemize}
\item We propose a unified spatial-temporal-semantic transformer that fuses surrounding cameras, adjacent frames, and semantic features. As the feature extraction backbone of our model, this module effectively merges spatial-temporal and semantic information to improve accuracy while reducing computational consumption.

\item We design a joint pose estimation network for surrounding cameras that integrates depth and extrinsic parameters, and this improves the network's interpretability while enhancing the accuracy of pose prediction.

\item We integrate the depth prediction world model into the loss function module and design a gradient-based curvature loss function, which accelerates the convergence and improves the quality of depth estimation.

\item We conduct comprehensive validation on two widely used datasets. The results demonstrate the superior capability of our algorithm to achieve SOTA performance in metric depth estimation for autonomous driving.

\end{itemize}

%% file: sec/2_relatedwork.tex
\vspace{-0.6em}
\section{Related Works}\vspace{-0.6em}
\label{sec:relatedwork}

\subsection{Scale-Ambiguous Depth Estimation}
Monocular depth estimation begins by predicting depth information from a single image~\cite{alhashim2018high,  ranftl2021vision, lee2019big,shi2023ega}. However, this approach typically builds a scale-ambiguous depth estimation world model rather than providing metric depth estimation due to the lack of scale information. 
Some studies~\cite{godard2017unsupervised, godard2019digging,guizilini20203d} use adjacent frames to calculate scale information, enabling continuous depth estimation in image sequences under reprojection error supervision. Building on this, ManyDepth~\cite{watson2021temporal} and~\cite{li2023learning} introduce outlier rejection methods for dynamic objects to reduce errors. SurroundDepth~\cite{wei2023surrounddepth} and~\cite{wang2023crafting} utilize additional data sources, such as depth priors and velocity, to improve results. 
However, due to the absence of reliable scale information, these methods are not applicable in scenarios requiring high metric precision. 

\vspace{-0.5em}
\subsection{Metric Depth Estimation}
MonoRec~\cite{wimbauer2021monorec} introduces a visual odometry system~\cite{yang2018deep} to provide relative pose estimation and sparse depth supervision, enabling scale-aware depth estimation. 
R3D3~\cite{schmied2023r3d3} proposes a multi-camera dense bundle adjustment method and a multi-camera co-visibility graph to compute accurate poses. 
However, modules in these works~\cite{wimbauer2021monorec,schmied2023r3d3} are not entirely differentiable, which increases the complexity of manually designed principles. 
VFDepth~\cite{kim2022self} constructs a volumetric feature representation as the backbone of the entire model, but the depth predictions are not clearly separated at the edge areas due to the lack of semantic information. 
SurroundDepth~\cite{wei2023surrounddepth} uses a cross-view transformer to exchange information, but the pose estimation network is computationally expensive due to the separate feature encoder. 
To achieve accurate metric depth estimation, the extrinsic parameters of a multi-camera setup and precise adjacent pose estimation are key factors.s 
Moreover, pose estimation solely derived from image data is difficult to convergence.
M$^2$-Depth~\cite{zou2024m} utilizes image pairs from a single camera to predict pose, but neglects the scale information provided by the extrinsic of surrounding cameras. 

\vspace{-0.5em}
\subsection{World Model Guided Metric Depth Estimation}

In recent years, with the rapid advancement of data and computational resources, remarkable progress has been made in semantic segmentation world models~\cite{kirillov2023segment,ravi2024sam,zhang2023faster,zhang2023mobilesamv2} and depth estimation world models~\cite{bhat2023zoedepth, yang2024depth, yang2024v2depth}. Some methods~\cite{godard2017unsupervised, godard2019digging,wei2023surrounddepth} that rely on reprojection error supervision are prone to degeneration in areas with semantically similar features. In contrast, M$^2$-Depth~\cite{zou2024m} and DepthAnything~\cite{yang2024depth, yang2024v2depth} incorporate semantic information into depth estimation tasks, significantly improving both the accuracy and plausibility of depth predictions. 
Although the depth estimated by world models~\cite{bhat2023zoedepth, yang2024depth, yang2024v2depth} is scale-ambiguous, the relative depth information they provide can guide the estimation of metric depth.

%% file: sec/3_method.tex
\section{Methodology}
\label{method}

\begin{figure*}[!htb]
\centering
\includegraphics[width=\linewidth]{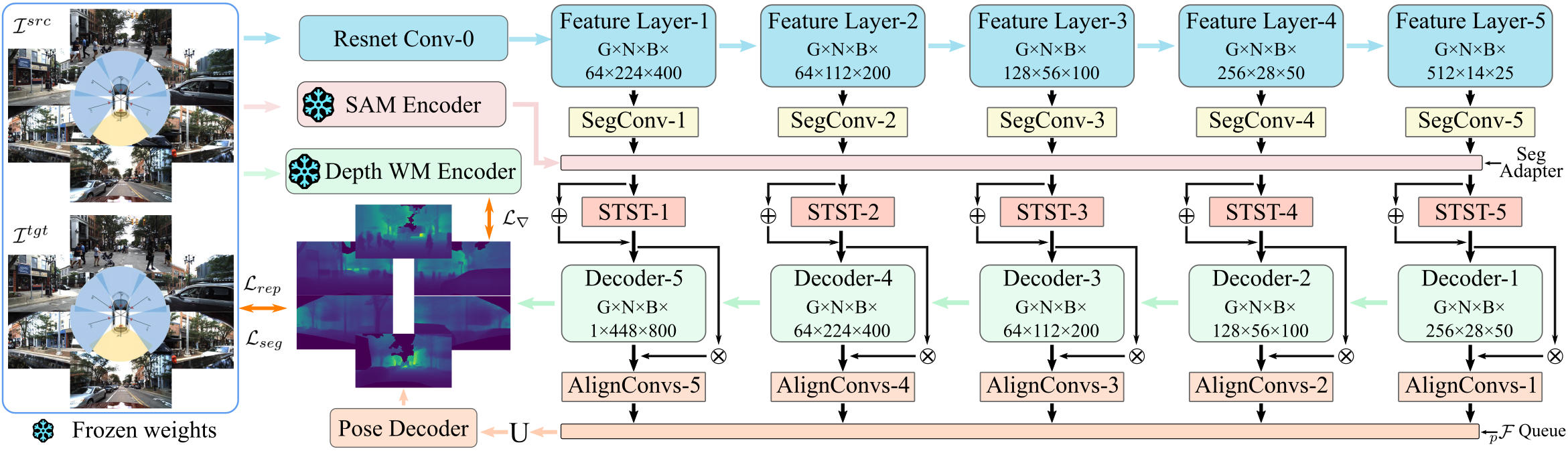}
\caption{Overview of our proposed framework. Our framework takes two adjacent surrounding frames as input. The core of the system consists of a feature extraction layer based on ResNet~\cite{he2016deep} and a depth decoder. Between the encoder and decoder in each layer, we use self-designed STST module to perform high-dimensional information fusion. Then the output from the depth decoder and the fused feature from the STST module are combined using the Hadamard product ($\otimes$), as the input of surrounding camera pose estimation network. The final output includes both a depth map and joint pose estimation. $\oplus$ represents element-wise addition of tensors. $\text{U}$ refers to the operation of concatenating the features from all layers.}
\label{fig:overview}
\vspace{-1.2em}
\end{figure*}

\subsection{Problem Formulation and Semantic Feature Adapter}
\noindent\textbf{Problem Formulation}. The input data in every iteration consists of $G$ ($G=2$ in our paper) adjacent surrounding frames, with each frame containing $N$ images. The $n$th image is defined as $\mathcal{I}_{n}^{3 \times H \times W}$, where $0 \leq n < N$. $H$ and $W$ denote the height and width of the image, respectively. We assume that the intrinsics $\pi_{n}$ of $n$th image and the extrinsic $\mathcal{E}_n$ between the $n$th camera and the base coordinate system are known. The $G$ frames are divided into source~\textit{(src)} frames and target~\textit{(tgt)} frames. Our framework aims to predict the inverse depth $\mathcal{D}_{n}^{1 \times H \times W}$ for each image $\mathcal{I}_{n}^{3 \times H \times W}$, as well as the pose $\mathcal{P}^{(G-1) \times 6}$ between the \textit{src} frames and the \textit{tgt} frames, represented by the 6-DOF axis-angle notation.

A brief overview of our framework is illustrated in~\myfigref{fig:overview}. We extract visual features from combined images $\mathcal{I}^{G \times N \times 3 \times H \times W}$ by ResNet~\cite{he2016deep}, and the output of the $L$-layers extracted features is denoted as $\mathcal{F}_{vis}=\{\mathcal{F}_{(l)}^{G \times N \times C_l \times H_l \times W_l},\, 0 \leq l < L\}$, where $C_l$, $H_l$, and $W_l$ represent the number of channels, feature height, and feature width at the $l$th layer, respectively. 
Meanwhile, semantic features $\mathcal{F}_{seg}$ are obtained from combined images $\mathcal{I}^{G \times N \times 3 \times H \times W}$ by the frozen SAM feature encoder~\cite{zhang2023mobilesamv2}.
Features $\mathcal{F}_{vis}$ and $\mathcal{F}_{seg}$ are then passed through a self-designed spatial-temporal-semantic transformer (STST) module (detailed in~\mysecref{sec:stst}) to fuse features. 
This fused feature, which aggregates temporal, spatial, and semantic information, serves as the input for both the depth decoder and the pose decoder, thereby avoiding redundant feature extraction~\cite{wei2023surrounddepth,zou2024m} from the raw images and making the network more lightweight. In~\mysecref{sec:pose}, we introduce the structure of the STST-enhanced joint pose estimation network. Following this, we detail the implementation of our loss function and network in~\mysecref{sec:loss}.

\noindent\textbf{Semantic Feature Adapter.}
We integrate semantic information pretrained by the semantic world model~\cite{zhang2023mobilesamv2,zhang2023faster} into the our network, constructing a fused feature to enhance depth estimation performance.
\( \mathcal{F}_{seg} \) denotes the semantic feature from MobileSAM~\cite{zhang2023mobilesamv2} with dimensions \( C_g \times H_g \times W_g \). 
For $l$th \textit{layer} visual features \( \mathcal{F}_{(l)}\), specific convolution and sampling operations are adapted to align with the dimensions of \( \mathcal{F}_{seg} \). This process can be expressed as
\begin{equation}
_{s}{\mathcal{F}_{(l)}} = DeConvs^{s}_{(l)}\left( Convs^{s}_{(l)}(\mathcal{F}_{(l)}) + \mathcal{F}_{seg} \right)
\end{equation}
where \( Convs^{s}_{(l)} \) and \( DeConvs^{s}_{(l)} \) represent the $l$th layers convolution and sampling operations designed to align the feature \( \mathcal{F}_{(l)} \) with the dimensions of \( \mathcal{F}_{seg} \).

\begin{figure}[t]
\centering
\includegraphics[width=0.85\linewidth]{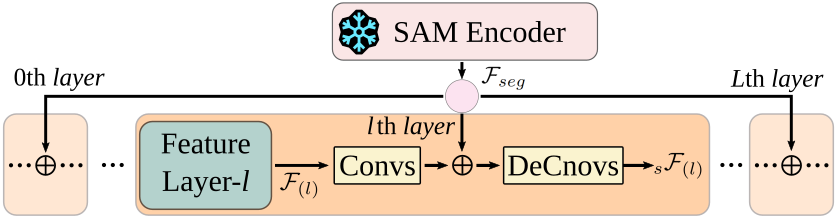}
\caption{Illustration of the semantic adapter network at the $l$th layer. All $L$ layers receive the same semantic feature $\mathcal{F}_{seg}$.}
\label{fig:mpe}
\vspace{-1.5em}
\end{figure}














The architecture we design integrates spatical-temporal features into the semantic layer, effectively avoiding the typical increase in parameters that arises when mapping from the semantic layer back to the feature layer. This approach can enhance depth estimation clarity in boundary area while maintaining computational efficiency.

\vspace{-0.7em}
\subsection{Spatial-Temporal-Semantic Transformer}
\label{sec:stst}
\noindent\textbf{Framework of Spatial-Temporal-Semantic Transformer.}
We build a unified spatial-temporal-semantic transformer architecture to fuse the features $\mathcal{F}_{vis}$ and $\mathcal{F}_{seg}$ extracted from images. For all $L$-layers, we use $L$ STST modules to facilitate feature exchange across surrounding cameras and adjacent frames.
The following section explains the construction process of STST in detail, using the $l$th layer as an example.

\begin{figure}[t]
\centering
\includegraphics[width=\linewidth]{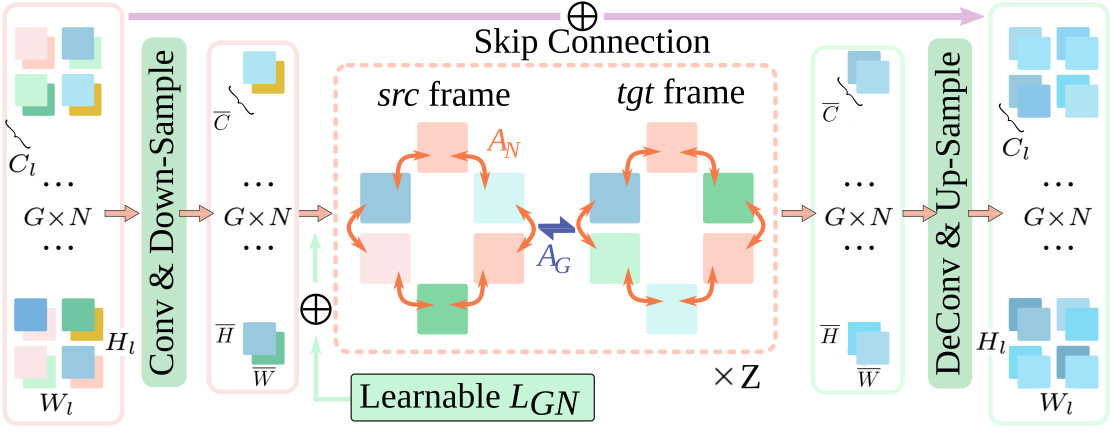}
\caption{Illustration of $l$th layer spatial-temporal-semantic transformer. Spatial attention is constructed using adjacent cameras, while temporal attention is built using adjacent frames. Before the attention calculation, the positions of the features are encoded in a learnable manner.}
\label{fig:stst}
\vspace{-1.5em}
\end{figure}

The overall structure of the $l$th STST is shown in~\myfigref{fig:stst}.
For the feature at the $l$th layer, $_{s}\mathcal{F}_{(l)}$, we output the feature $_{sts}\mathcal{F}_{(l)}$ with spatial-temporal-semantic information, which retains the same dimension.
To reduce computational cost, we first down-sample the feature $_{s}\mathcal{F}_{(l)}^{G \times N \times C_l \times H_l \times W_l}$ with different down-sampling ratios for different layers, such that the shape of down-sampled feature matches the unified input shape of STST $G \times N \times \overline{C} \times \overline{H} \times \overline{W} $. 
Note that 
 $\overline{C}$, $\overline{H}$ and $\overline{W}$ denote the fixed number of channels, feature width, and feature height accepted by STST. Further details are provided in our code.

Then, by applying self-designed surrounding camera attention and adjacent frame attention detailed in~\mysecref{sec:spattention}, followed by upsampling to restore the shape, we obtain the desired output $_{sts}{\mathcal{F}}_{(l)}^{G \times N \times C_l \times H_l \times W_l}$. Additionally, we use skip connections between the encoder and decoder of the $l$th layer to retain gradient information.

\noindent\textbf{Spatial Surrounding Camera Attention and Temporal Adjacent Frame Attention.}\label{sec:spattention}
By leveraging features from overlapping regions and extrinsic parameters between cameras, we design a spatial attention mechanism for neighboring cameras to inference the metric scale information.
Unlike SurroundDepth~\cite{wei2023surrounddepth}, we reduce computational load by only computing interactions with clockwise neighbors. 
Additionally, we introduce a temporal attention mechanism for inter-frame feature matching to improve joint pose estimation accuracy. The cross-attention module is shown in~\myfigref{fig:stst}, the input feature is 
$
_{s}\mathcal{F}_{(l)}^{G \times N \times \overline{C}  \times \overline{H}  \times \overline{W}} $
and the output feature with same dimension is 
$
_{sts}\mathcal{F}_{(l)}^{G \times N \times \overline{C}  \times \overline{H}  \times \overline{W} }.
$

In this module, we incorporate learnable positional encoding $L^{G\times N\times 1 \times 1 \times 1}_{GN}$ in both the $G$ and $N$ dimensions. The feature with element-wise positional encoding is represented as
\begin{equation}
_{s}\mathcal{F}_{(l)}^{G \times N \times  \overline{C}  \times  \overline{H}  \times  \overline{W} } ={_{s}\mathcal{F}}_{(l)}^{G \times N \times  \overline{C}  \times  \overline{H}  \times  \overline{W}} \oplus bc(L_{GN})
\end{equation}
where $bc$ refers to using the broadcasting
mechanism to expand the dimensions of $L_{GN}$, transforming it into $L^{G\times N\times \overline{C}  \times  \overline{H}  \times  \overline{W}}_{GN}$.
Based on this, we use a linear layer $\left \langle\mathbf{W_{in}},\, \mathbf{b_{in}}\right \rangle$ to map this feature to 
$^{proj}\mathcal{F}_{(l)} = \mathbf{W_{in}} \cdot Flatten(\mathcal{F}_{(l)}) + \mathbf{b_{in}}
$, where $Flatten$ refers to the operation of converting a tensor into a one-dimensional array.
Then, for the \(n\)th camera, we compute the cross-attention score \(A_N\) with its adjacent camera. The query, the key and value are defined as
\begin{equation}
\begin{aligned}
Q_n &= _{\quad\, s}^{proj}\mathcal{F}_{(l)}[\,:\,, \, n,\, ...] \\
K_{n+1}, \, V_{n+1} &= _{\quad\, s}^{proj}\mathcal{F}_{(l)}[\,:\,, \, (n+1) \% N,\, ...]
\end{aligned}
\end{equation}
The computation of spatial attention \(A_N\) is given by
\begin{equation}
\label{eq:attn_N}
A_N = softmax\left(\frac{Q_{n}K_{n+1}^{\top}}{\sqrt{d_{n+1}}}\right) V_{n+1}
\end{equation}

For the $src$ frame, we compute the cross-attention score \(A_G\) with its adjacent $tgt$ frame. The query, key and value are defined as defined as
\begin{equation}
\begin{aligned}
Q_{src} &=_{\quad\, s}^{proj}\mathcal{F}_{(l)}[src,\,...] \\
K_{tgt},\, V_{tgt} &= _{\quad\, s}^{proj}\mathcal{F}_{(l)}[tgt,\,...]
\end{aligned}
\end{equation}
The computation of temporal attention \(A_G\) is given by
\begin{equation}
\label{eq:attn_G}
A_G = softmax\left(\frac{Q_{src}K_{tgt}^{\top}}{\sqrt{d_{tgt}}}\right) V_{tgt}
\end{equation}

We then combine the computed features \(_{attn}\mathcal{F}_{(l)} = \langle A_N, A_G \rangle\) and apply layer normalization followed by linear transformation $\left \langle\mathbf{W_{0}},\, \mathbf{b_{0}}\right \rangle , \left \langle\mathbf{W_{out}},\, \mathbf{b_{out}}\right \rangle $ to project it back to the original dimension:
\begin{equation}
\begin{aligned}
_{norm}\mathcal{F}_{(l)} &= LayerNorm(_{attn}\mathcal{F}_{(l)}) \\
_{sts}\mathcal{F}_{(l)} &= \mathbf{W_{out}} \cdot ReLU (\\ 
&\quad 
\mathbf{W_{0}} \cdot _{norm}\mathcal{F}_{(l)} + \mathbf{b_{0}}) + \mathbf{b_{out}}
\end{aligned}
\end{equation}

To reduce the computational cost of the attention module, we employ a convolutional projection in practice to decrease the channel dimension $C$. For simplicity, this modification is omitted in the theoretical derivation above. The implementation details are available in our source code.

\vspace{-0.7em}
\subsection{Multi-Camera Enhanced Pose Estimation}
\label{sec:pose}

Unlike most image-based methods~\cite{zou2024m,wei2023surrounddepth,godard2019digging}, we believe that the results of depth map prediction can provide geometric guidance for pose estimation. Furthermore, the coarse-to-fine pose refinement approach used in traditional methods can still be applied in deep neural network. 
Therefore, we combine the features \( _{sts}\mathcal{F}_{(l)} \) extracted from the \( l \)th layer of the STST backbone with the depth prediction \( _{d}\mathcal{F}_{(l)} \) from the \( l \)th depth decoder, and then utilize a multi-stage convolutional network to integrate these features.

\begin{figure}[t]
\centering
\includegraphics[width=\linewidth]{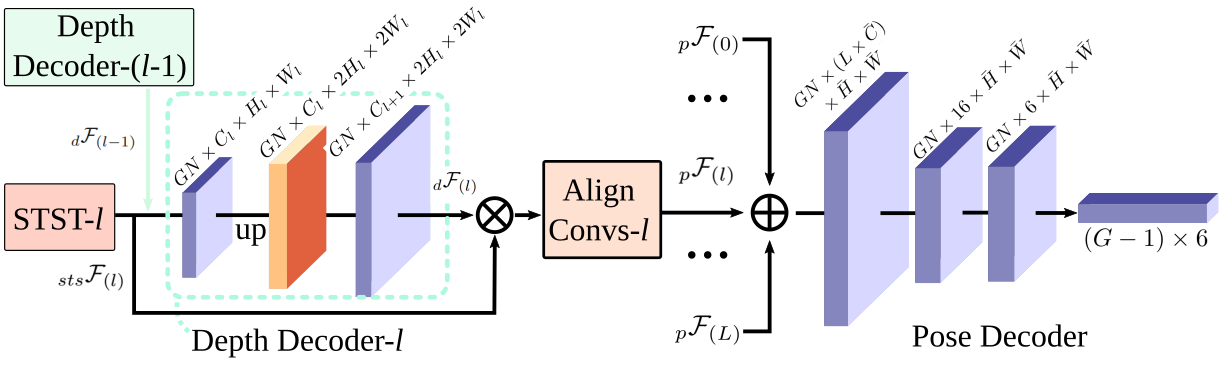}
\caption{Illustration of the pose feature fusion network at the $l$th layer and the pose decoder. By overlaying the predicted depth features onto the image features, depth-guided information is provided for the global pose estimation of the surrounding camera.}
\label{fig:mpe}
\vspace{-1.5em}
\end{figure}

At the $l$th layer, we unitize scale-adaptive compression as:
\begin{equation}
_{p}\mathcal{F}_{(l)} = Convs^{d}_{(l)}(_{sts}\mathcal{F}_{(l)} \otimes \, _{d}\mathcal{F}_{(l)})
\end{equation}
where $Convs^{d}_{(l)}$ denotes the convolution and sampling kernel. The concatenated features of all $L$ layers will be
$
_{p}\mathcal{F} = \cup_{l=1}^L {(_{p}\mathcal{F}_{(l)})}.
$
After aligning the feature dimensions, a lightweight pose decoder with convolution and mean operations is employed to generate the final joint pose estimation $\mathcal{P}_{s \to t}$. The overview of pose decoder and $l$th depth decoder are shown in ~\myfigref{fig:mpe}.

Upon obtaining the prediction of global pose transformation $\mathcal{P}_{s \to t}$. The extrinsic parameters are utilized to convert the global transformation into individual pose transformations within the coordinate system of each camera by
\begin{equation}
\mathcal{P}_{n} = \mathcal{E}_n^{-1} \cdot \mathcal{P}_{s \to t} \cdot \mathcal{E}_n
\end{equation}
where $\mathcal{P}_{n}$ denotes the pose transformation of the 
$n$th camera coordinate system, while and $\mathcal{E}_n$ represents the pre-calibrated extrinsic parameters between $n$th camera and the base coordinate. Once the poses for all cameras are calculated, the image-level loss is computed to provide supervision information.


\begin{figure*}[htbp]
\centering
\includegraphics[width=0.99\linewidth]{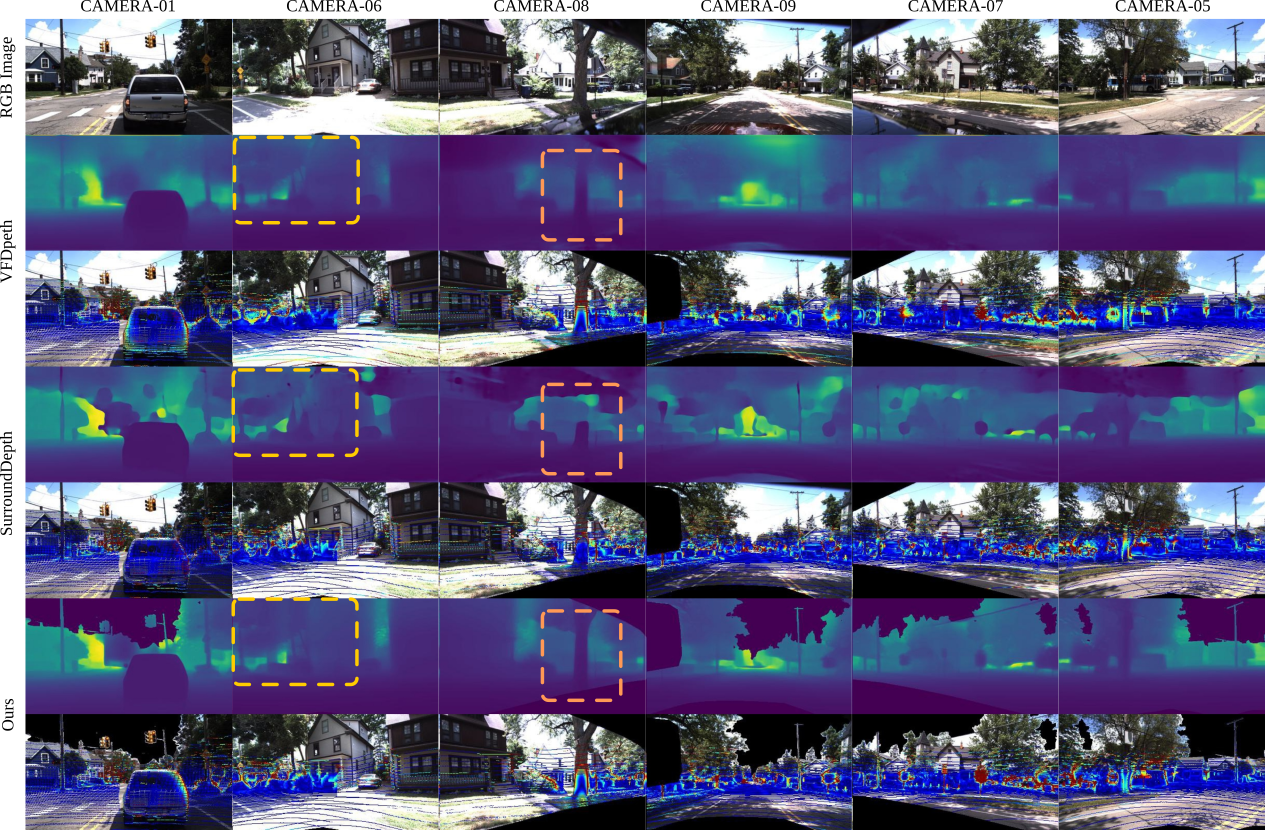}
\caption{Illustration of the metric depth estimation result on DDAD~\cite{guizilini20203d}. For each framework, we present the depth estimation result and the visilization of metric \textit{Abs.Rel}. The error distribution is same as~\myfigref{fig:firstdemo}. The boxes highlight some significant comparison details.}
\label{fig:ddad-comp}
\vspace{-1.5em}
\end{figure*}


\vspace{-0.7em}
\subsection{Implementation Details}
\label{sec:loss}
In our framework, we predict a full-size depth map, and then reshape it to the $k$th level ($0<k\leq K, K=3$ in our experiments) in a pyramid structure. The loss is calculated across all $K$ levels.

\noindent\textbf{Sparse Depth Loss.}
Previous methods~\cite{zou2024m,wei2023surrounddepth,kim2022self,schmied2023r3d3} rely on LiDAR point clouds captured at a single moment within an image frame for depth estimation validation. To ensure fair comparison, we also utilize the same data for metric validation and sparse supervision.

To compute the loss between the ground truth depth image and the predicted depth image, we select depth values within a specific range $D_{min} \to D_{max}$, and calculate the absolute error by
\begin{equation}
\label{eq:depthloss}
L_{d} =  \frac{1}{D_{max}}\sum_{k=0}^{K} \|\frac{1}{\mathcal{D}_{(k)}^{gt}} - \frac{1}{\mathcal{D}_{(k)}^{pred}} \|_1
\end{equation}
where \(\mathcal{D}^{gt}_{(k)}\) and \(\mathcal{D}^{pred}_{(k)}\) represent the ground truth and predicted inverse depth in $k$th level, respectively. Note that, this adjusts the data distribution of inverse depth resulting in a more uniform distribution that is more conducive for supervision.

\noindent\textbf{Curvature Loss.} In our experiments, first-order gradient-based methods~\cite{wei2023surrounddepth,zou2024m,godard2019digging} perform poorly in predicting depth details. We find that although the depth estimation world model~\cite{bhat2023zoedepth, yang2024depth, yang2024v2depth} is not accurate in scale, its depth prediction distribution closely resembles the real distribution. Therefore, we use curvature (i.e., second-order gradients) to measure the loss between the predicted depth $\mathcal{D}^{pred}$ and the depth estimation $\mathcal{D}^{wm}$ from MobileSAM~\cite{yang2024v2depth}.

We define a $t$-step ($0 < t \le T, T=3$ in our experiments) depth gradient operator for coordinate $x,y$ as
\begin{equation}
\nabla^{(t)} \left \{  \mathcal{D}(y, x) \right \} = \mathcal{D}(y, x) - \mathcal{D}(y + t, x + t)
\end{equation}
which can be applied to calculate curvature of the depth by
\begin{equation}
C^{(t)} \mathcal{D}(y, x) = \nabla^{(t)} \left\{ \nabla^{(t)}  \left \{ \mathcal{D}(y, x) \right \}\right \}
\end{equation}
Then, the curvature loss $\mathcal{L_{\nabla}}$ is defined as
\begin{equation}
\mathcal{L_{\nabla}} = \sum_{k = 0}^{K}\sum_{t = 1}^{T}  \| C^{(t)}\mathcal{D}_{(k)}^{pred}  \\ - C^{(t)}D_{(k)}^{wm}  \|_1
\end{equation}

Our experiments show that this curvature loss significantly improves the performance of depth prediction in edge details. More details can be found in~\mysecref{sec:abla}.

%

\noindent\textbf{Reprojection Loss and Semantic Loss.}
We predict the inverse depth $\mathcal{D}^{src}_{(n)(k)}$ of $n$th image in $k$th level and relative pose $\mathcal{P}_{n}$. The operator $\Lambda_{(n)}^{(k)}$ is defined as reprojecting image $\mathcal{I}^{src}_{(n)(k)}$ to $0$th layer $\mathcal{I}^{tgt}_{(n)(0)}$ by combining with sky and optional vehicle body occlusion mask ($\mathcal{M}^{src}_{(n)(k)}$, $\mathcal{M}^{tgt}_{(n)(k)}$):
\begin{equation}
\Lambda_{(n)}^{(k)} = \pi_{(n)(0)}^{tgt} \mathcal{P}_{n} (\pi_{(n)(k)}^{src})^{-1} (\mathcal{M}^{src}_{(n)(k)}\!\odot \!\mathcal{D}_{(n)(k)}^{src}\! \odot \!\mathcal{I}_{(n)(k)}^{src})
\end{equation}
where $\pi^{src}_{(n)(k)}$ and $\pi^{tgt}_{(n)(0)}$ are the intrinsic of $\mathcal{I}^{src}_{(n)(k)}$ and $\mathcal{I}^{tgt}_{(n)(0)}$, respectively. $\odot$ denotes the Hadamard product applied element-wisely to the predicted depth.
We calculate reprojection loss $\mathcal{L}_{rep}$ via L1 distance and SSIM distance~\cite{wang2004image} between $N$ $src$ frames $\mathcal{I}^{src}_{(n)(k)}$ and $N$ $tgt$ frames $^{M}\mathcal{I}^{tgt}_{(n)(0)} \leftarrow M^{tgt}_{(n)(0)}\odot \mathcal{I}^{tgt}_{(n)(0)}$. 
$\mathcal{L}_{rep}$ is defined as
\begin{equation}
\begin{aligned}
\mathcal{L}_{rep} &= \sum_{l=0}^{K}\sum_{n  = 0}^{N} \Bigg( \lambda_{l1}\|^{M}\mathcal{I}^{tgt}_{(n)(0)} -\Lambda^{(k)}_{(n)}( \mathcal{I}^{src}_{(n)(k)})\|_1 \\
&+ (1 -\lambda_{l1}) \, \text{SSIM}(^{M}\mathcal{I}^{tgt}_{(n)(0)},\, \Lambda^{(k)}_{(n)}(\mathcal{I}^{src}_{(n)(k)}))\Bigg)
\end{aligned}
\end{equation}
where $\lambda_{l1}=0.2$ in our experiments.

As for semantic loss, a pre-trained semantic world model $Seg$~\cite{zhang2023mobilesamv2} is used to extract semantic features $\mathcal{I}^{tgt}_{(0)}$. The goal is to align these features in a high-dimensional space during the training process. The alignment is supervised by minimizing
\begin{equation}
\mathcal{L}_{seg} = \sum_{n  = 0}^{N} \|Seg\left(\mathcal{I}^{tgt}_{(n)(0)}\right) - Seg\left(\Lambda^{(k)}_{(n)}(\mathcal{I}^{src}_{(n)(k)})\right)\|_1
\end{equation}

\noindent\textbf{Total Loss.}
Four losses are incorporated in our experiment. Since it is quite challenging to balance the weights of these four losses, we align the scales of all losses to the depth loss $\mathcal{L}_{d}$ and then apply the weights $\lambda_1=\lambda_2=0.5,\, \lambda_3=\lambda_4=3$. The total loss $\mathcal{L}_{all}$ is calculated by
\begin{equation}
\mathcal{L}_{all} = \lambda_1 \mathcal{L}_{d} + \lambda_2 \frac{|\mathcal{L}_{d}|\mathcal{L}_{\nabla}}{|\mathcal{L}_{\nabla}|} + \lambda_3 \frac{|\mathcal{L}_{d}|\mathcal{L}_{rep}}{|\mathcal{L}_{rep}|} + \lambda_4 \frac{|\mathcal{L}_{d}|\mathcal{L}_{seg}}{|\mathcal{L}_{seg}|}.
\end{equation}

%% file: sec/4_experiments.tex
\section{Experiments}
\label{exp}
In this section, we first introduce the experimental setup in \mysecref{sec:expsetup}, including datasets, evaluation metrics, baselines, and parameter settings, etc. Then in \mysecref{sec:results}, we mainly compare the frameworks performance with baselines. \mysecref{sec:abla} shows ablation experiments of the proposed framework. \mysecref{sec:gputime} shows GPU usage and inference time experiments of the proposed framework.


\begin{table*}[htb]
\caption{Quantitative comparison results of our method with the baselines on public datasets. $*$ represents the result from original paper as no available code. The results ranked from best to worst are highlighted as \colorbox[HTML]{c0e2ca}{first}, \colorbox[HTML]{fff5b3}{second}, and \colorbox[HTML]{ffd9b3}{third}.}
\label{tab:withbaseline}
\centering
\resizebox{0.8\linewidth}{!}{%
\begin{tabular}{@{}ccccccccccc@{}}
\toprule
      &Image Size& Supervision & Dataset                  & \textit{Abs.Rel.}↓ & \textit{Sq.Rel.}↓ & \textit{RMSE}↓ & \textit{RMSE log}↓  & $\delta<1.25$↑ & $\delta<1.25^2$↑ & $\delta<1.25^3$↑ \\ \midrule
R3D3~\cite{schmied2023r3d3}  & 640 $\times$ 384 & Semi&\multirow{5}{*}{DDAD~\cite{guizilini20203d}}      &  0.392      &   3.824   &    14.435     &                                     0.447 &                                                       0.482 & 0.623        &                                                         0.720 \\
VFDepth~\cite{kim2022self} & 640 $\times$ 384&Self&                         &     \colorthird0.259     &  \colorthird3.340       &   12.934   &        \colorthird0.362 &    0.693                                  &                                                        0.795 &                                                        0.892 \\
SurroundDepth~\cite{wei2023surrounddepth}   &                                                        640 $\times$ 384 & Semi& &                         0.273 &    3.540           &    \colorthird12.651     &     0.372 &     \colorthird0.790    &                                      \colorthird0.883                         &                                                         \colorthird0.930                                                                                                       
\\
M$^2$Depth$^*$~\cite{zou2024m}      &640 $\times$ 384 & Semi&                        &                                                         \colorsecond0.182 &      \colorsecond2.920    &    \colorsecond11.963     & \colorsecond0.299     &  \colorsecond0.756       &   \colorsecond0.897                                   &                                                        \colorsecond0.947 \\
Ours &  800 $\times$ 448& Semi&     &     \colorfirst0.167              &     \colorfirst2.686&     \colorfirst11.407    &  \colorfirst0.283    &      \colorfirst0.792   &     \colorfirst0.904        &    \colorfirst0.953      \\ \midrule
R3D3~\cite{schmied2023r3d3}   & 768 $\times$ 448&Semi& \multirow{5}{*}{nuScenes~\cite{caesar2020nuscenes}} &                  0.368&5.985 &   8.547   &     0.409    &                                     0.639 &                                                        0.742 &                                                        0.832 \\
VFDepth~\cite{kim2022self} &640 $\times$ 352& Self&                         &    \colorthird0.284      &  \colorthird4.892       & \colorthird6.982     &        0.385 &    0.640                                  &                                                        \colorthird0.773 &                                                        \colorthird0.873 \\
SurroundDepth~\cite{wei2023surrounddepth}      & 640 $\times$ 352 &Semi&                        &   0.309       &  5.232       &  7.106    &        \colorthird0.342 &                     \colorthird0.672                 &                                                        0.741 &                                                        0.869
\\
M$^2$-Depth$^*$~\cite{zou2024m}      &640 $\times$ 352 &  Semi&                     &                                                         \colorsecond0.259 &      \colorsecond4.599    &    \colorsecond6.898     & \colorsecond0.332     & \colorsecond 0.734       &   \colorsecond0.871                                   &                                                        \colorsecond0.928  \\
Ours &     800 $\times$ 448    & Semi&                &      \colorfirst0.197    &    \colorfirst2.624     &   \colorfirst6.094   &   \colorfirst0.297      &   \colorfirst0.789                                   & \colorfirst0.903                                                        &                                                        \colorfirst0.946 \\ \bottomrule
\end{tabular}%
}
\vspace{-1.5em}
\end{table*}

\vspace{-0.5em}
\subsection{Environmental Setup}
\label{sec:expsetup}

\noindent\textbf{Baselines and Metrics.}
We compare our results with R3D3~\cite{schmied2023r3d3}, VFDepth~\cite{kim2022self}, scale-aware SurroundDepth~\cite{wei2023surrounddepth}, and M$^2$-Depth~\cite{zou2024m}. 
It is pertinent to note that both SurroundDepth~\cite{wei2023surrounddepth} and M$^2$-Depth~\cite{zou2024m} utilize the structure-from-motion (SfM) method to generate sparse depth for training supervision, contradicting their claim of being self-supervised in their papers. Therefore, in the comparative experiments of this paper, we categorize them as semi-supervised methods.
Following prior works~\cite{guizilini2022full, zou2024m, wei2023surrounddepth}, the evaluation metrics we used are $Abs.Rel.$, $Sq.Rel.$, $RMSE$, $RMSE\,log$, and $\delta$.

\noindent\textbf{Datasets.}
We train and evaluate our framework on two public datasets, including DDAD~\cite{guizilini20203d} and nuScenes~\cite{caesar2020nuscenes}. For the DDAD~\cite{guizilini20203d} dataset, we first crop the image from the top-left corner to a size of 1936$\times$1084, and then resize it to 800$\times$448. For the nuScenes~\cite{caesar2020nuscenes} dataset, we directly resize the images to 800$\times$448. To the best of our knowledge, the image sizes employed in our work are the largest, which helps improve the extraction of detailed features and their subsequent refinement. 

The maximum evaluation depth $D_{max}$ is set to 200 meters, averaged across all cameras in the DDAD~\cite{guizilini20203d} dataset, and 80 meters in the nuScenes~\cite{caesar2020nuscenes} dataset, consistent with the baselines. For the nuScenes dataset~\cite{caesar2020nuscenes}, we select 300 scenes featuring aggressive driving scenarios to avoid overfitting and reduce the computational cost of experiments.

\noindent\textbf{Training.}
We implement our approach using PyTorch 
and train the model with the Adam optimizer at a learning rate of 10$^{-4}$. The training runs for 100 epochs across all datasets, with the first 5 epochs dedicated to warm-up for the learning rate, which is then decayed using a cosine schedule. 
To ensure a fair comparison, we use a 34-layer ResNet~\cite{he2016deep} as the backbone.
For semantic feature extraction, we utilize the frozen SAM encoder provided by MobileSAM~\cite{zhang2023mobilesamv2} to reduce GPU usage. We use SegFormer~\cite{xie2021segformer} to generate the sky mask, removing interference from the sky in outdoor scenes.
For the DDAD dataset~\cite{guizilini20203d}, we also eliminate the vehicle occlusion areas using the corresponding mask data from~\cite{wei2023surrounddepth}.  Our experiments are conducted on 8 NVIDIA H100 80GB HBM3 GPUs, with 50 hours of training on the nuScenes~\cite{caesar2020nuscenes} dataset and 16 hours on the DDAD~\cite{guizilini20203d} dataset. Additional details can be found in our released code.

\vspace{-0.5em}
\subsection{Qualitative and quantitative analysis}
\label{sec:results}


\noindent\textbf{Comparison with Baselines}. We begin by presenting a qualitative comparison of different baselines and our method on the DDAD~\cite{guizilini20203d}, as shown in~\myfigref{fig:ddad-comp}.
It is clear that our method yields clearer image details and sharper boundaries. And the visilization of  \textit{Abs.Rel.} demostrate the quality of depth estimation is also superior. The quantitative comparison on the DDAD~\cite{guizilini20203d} and nuScenes~\cite{caesar2020nuscenes} datasets is shown in~\mytabref{tab:withbaseline}.

\noindent\textbf{Per-Camera Evaluation}. We also do per-camera evaluation in~\mytabref{tab:percamera}. M$^{2}$-Depth~\cite{zou2024m} estimate the pose by the front camera, so the metric ~\textit{Abs.Sql} on front camera is the best. But our framework can achieve a error balance across surrounding cameras by global feature fusion.

\begin{table}[!htb]
\vspace{-0.5em}
\caption{Per-camera \textit{Abs.Sql} evaluation on DDAD~\cite{guizilini20203d}.}
\label{tab:percamera}
\resizebox{\linewidth}{!}{%
\begin{tabular}{@{}cccccccc@{}}
\toprule
\multirow{2}{*}{Methods} & \multicolumn{6}{c}{\textit{Abs.Sql}}                         &      \\
                         & CAM-01 & CAM-05 & CAM-06 & CAM-07 & CAM-08 & CAM-09 & Avg. \\ \midrule
R3D3~\cite{schmied2023r3d3}                    &0.375& 0.428&0.417&0.388&0.381&0.363    &0.392      \\
VFDepth~\cite{kim2022self}                  &  \colorthird0.240&\colorthird0.260&0.277&\colorthird0.251&\colorthird0.253&\colorthird0.273&  \colorthird0.259     \\
SurroundDepth~\cite{wei2023surrounddepth}                       &  0.263&0.279&\colorthird0.258&0.254&0.292&0.291&  0.273      \\
M$^2$-Depth~\cite{zou2024m}               & \colorfirst0.146       &    \colorsecond0.182    &        \colorsecond0.200&    \colorsecond0.198    &   \colorsecond0.203      &     \colorsecond0.169   &      \colorsecond0.183 \\
Ours             & \colorsecond0.161&\colorfirst0.172&\colorfirst0.169&\colorfirst0.171&\colorfirst0.168&\colorfirst0.167&       \colorfirst0.155     \\ \bottomrule
\end{tabular}%
}
\vspace{-1.5em}
\end{table}

\vspace{-0.5em}
\subsection{Ablation study}
\label{sec:abla}

\begin{figure*}[!htb]
\centering
\includegraphics[width=1\linewidth]{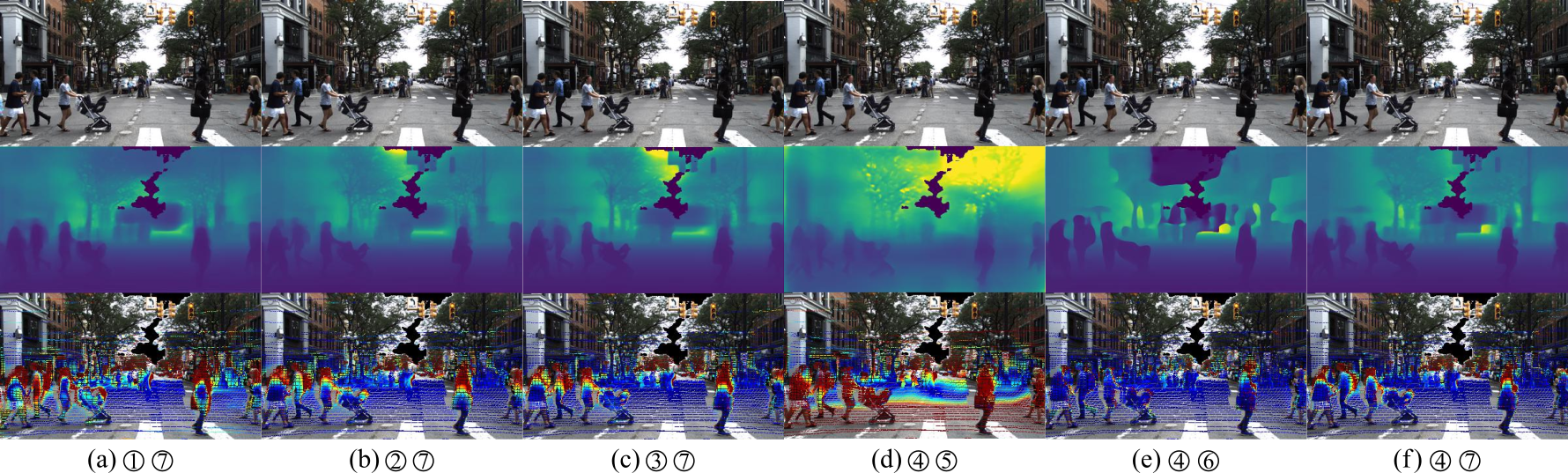}
\caption{Visualizations of the ablation experiments on metric depth estimation results for various modules, with parameter configurations and combinations corresponding to the IDs provided in~\mytabref{tab:abla-module} and~\mytabref{tab:abla-loss}.}
\label{fig:abla-module}
\vspace{-1.5em}
\end{figure*}

\begin{figure}[!htb]
\centering\vspace{-1em}
\includegraphics[width=\linewidth]{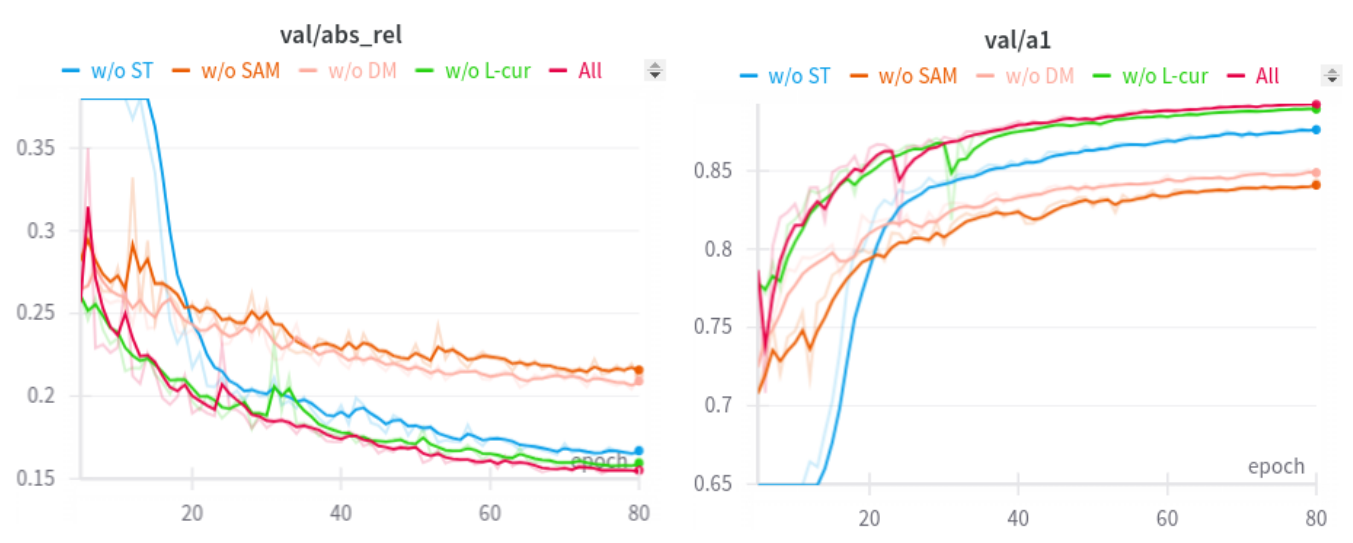}
\caption{Visualization of training iterations on DDAD~\cite{guizilini20203d} dataset. It can be observed that as we progressively incorporate the proposed modules, the overall model's metric accuracy and convergence speed improve steadily. $val/abs\_rel$ and $val/a1$ represent $Abs.Rel$ and $\delta < 1.25$, respectively.}
\label{fig:iteration}
\vspace{-1em}
\end{figure}

\noindent\textbf{Ablation Study on Module Contributions.} We systematically evaluate the impact of various modules and losses on the final performance, with a particular focus on the spatial-temporal attention (ST) mechanism, SAM feature integration (SAM), depth-enhanced motion estimation networks, the sparse depth loss $\mathcal{L}_d$, and curvature loss $\mathcal{L}_{\nabla}$. The quantitative effects of different module and loss combinations are presented in~\mytabref{tab:abla-module} and~\mytabref{tab:abla-loss}, while qualitative results are illustrated in~\myfigref{fig:abla-module}.
\myfigref{fig:iteration} shows the convergence of our different modules and their final metrics on the validation set.
The results demonstrate that the key components of the framework significantly enhance overall performance, both in qualitative and quantitative evaluations.
It is worth noting that with the introduction of our curvature loss $\mathcal{L}_{\nabla}$, although there is no significant improvement in the metrics in~\mytabref{tab:abla-loss}, there is a substantial enhancement in the smoothness and semantic consistency of the depth map shown in~\myfigref{fig:abla-module}. This is because the depth loss is inherently sparse, constraining only a minimal number of regions in the depth predictions.


\begin{table}[!htb]
\vspace{-0.5em}
\caption{Qualitative comparison results on DDAD (evaluate depth: 80\,m) of the three modules: spatial-temporal attention (ST), SAM feature (SAM), depth-enhanced motion estimation (DM).}
\centering
\label{tab:abla-module}
\resizebox{\linewidth}{!}{%
\begin{tabular}{@{}cccccccccccc@{}}
\toprule
                  \multicolumn{1}{c}{ID} & ST& SAM & DM & \textit{Abs.Rel.}↓ & \textit{Sq.Rel.}↓ & \textit{RMSE}↓ & \textit{RMSE log}↓& $\delta<1.25^3$↑ \\ \midrule
                             \ding{172}    &           \ding{55}     &   \ding{55}   &   \ding{55}       &      0.223    &  5.726       &  6.897    &                                                        0.329 &                                                        0.923 \\
                             \ding{173}    &           \ding{51}     &   \ding{55}   &   \ding{55}       &   \colorthird0.206       &   \colorsecond4.513      &   \colorthird6.708   &                                                        \colorthird0.310 &                                                        \colorthird0.959 \\
                             \ding{174}    &      \ding{51}          &   \ding{51}  &  \ding{55}       &      \colorsecond0.204    &    \colorthird4.573     &   \colorsecond6.494   &                                                        \colorsecond0.298 &                                                        \colorsecond0.961 \\
                          \ding{175}   &    \ding{51}               &   \ding{51}  &  \ding{51}     &                   \colorfirst0.155 &  \colorfirst3.307  &  \colorfirst5.934 &  \colorfirst0.276 &\colorfirst0.968          
                         \\\bottomrule
\end{tabular}%
}
\vspace{-1.5em}
\end{table}

\begin{table}[!htb]
\vspace{-0.5em}
\caption{Qualitative comparison results on DDAD (evaluate depth: 80\,m) of the two losses: depth loss $\mathcal{L}_d$ and curvature loss $\mathcal{L}_\nabla$.}
\centering
\label{tab:abla-loss}
\resizebox{0.95\linewidth}{!}{%
\begin{tabular}{@{}cccccccccccc@{}}
\toprule
                  \multicolumn{1}{c}{ID}&$\mathcal{L}_d$  & $\mathcal{L}_{\nabla}$ & \textit{Abs.Rel.}↓ & \textit{Sq.Rel.}↓ & \textit{RMSE}↓ & \textit{RMSE log}↓   & $\delta<1.25^3$↑ \\ \midrule
                        \ding{176}    &     \ding{55}  &   \ding{51}      &  \colorthird3.319       &        \colorthird165.719 &                                          \colorthird36.792 &                                                        \colorthird1.366 &                                                         \colorthird0.323
\\
                    \ding{177}    &   \ding{51}  &    \ding{55}            &  \colorsecond0.163    &        \colorsecond3.310 &                                     \colorsecond6.337 &                                                        \colorsecond0.279 &                                                     \colorsecond0.967
\\
              \ding{178}    & \ding{51}  &   \ding{51}     &                   \colorfirst0.155 &  \colorfirst3.307  &  \colorfirst5.934 &  \colorfirst0.276& \colorfirst0.968          
                         \\\bottomrule
\end{tabular}%
}
\vspace{-1.5em}
\end{table}



\begin{figure}[!htb]
\centering
\includegraphics[width=0.9\linewidth]{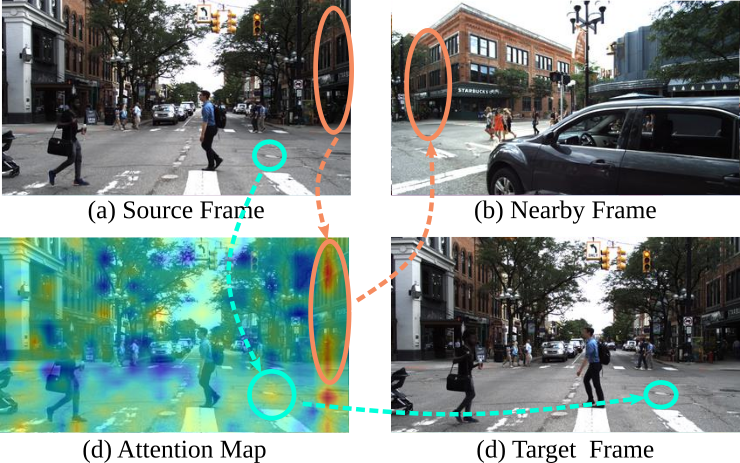}
\caption{Visualization of attention map on DDAD~\cite{guizilini20203d} dataset. It can be observed that the attention module highlights certain key areas, which contribute to the estimation of scale information and the pose transformation estimation between adjacent frames.}
\label{fig:attn}
\vspace{-0.8em}
\end{figure}

\noindent\textbf{Attention Map Visualizations.} We present visualizations of attention maps for both spatial surrounding camera attention and temporal frame attention in~\myfigref{fig:attn}. We observe that the spatial surrounding camera attention predominantly focuses on overlapping regions (highlighted in orange), effectively capturing matching relationships between adjacent surrounding cameras. Meanwhile, the temporal frame attention emphasizes regions with more distinctive features (highlighted in blue), which aids in robust frame-to-frame matching. These visualizations demonstrate the effectiveness of our attention modules in capturing spatial and temporal relationships.

\vspace{-0.5em}
\subsection{GPU Usage and Time Consumption}
\label{sec:gputime}
We compare the GPU usage and time consumption of our models with other baselines, as shown in~\mytabref{tab:gputime}. Note that we use the semantic world model~\cite{zhang2023mobilesamv2} to obtain \textit{online} semantic features, which we have taken into consideration. It is evident that our model demonstrates improvements in time efficiency. Although M$^2$-Depth~\cite{zou2024m} claims to also utilize semantic features, we cannot verify whether they use the offline semantic world model to reduce the GPU usage due to the lack of source code.

\begin{table}[!htb]
\vspace{-0.5em}
\caption{Comparison of GPU usage and inference time of one frame. Best results are underlined.}
\centering
\label{tab:gputime}
\resizebox{0.7\linewidth}{!}{%
\begin{tabular}{@{}ccc@{}}
\toprule
Models & GPU (Mb) & Inference Time (milliseconds) \\ \midrule
SurroundDepth~\cite{wei2023surrounddepth}      &                        8732 &  248   \\ 
M$^2$-Depth$^*$~\cite{zou2024m}      &     \underline{5546}     &                    295 \\
Ours &    9137      &                  \underline{232}   \\ \bottomrule
\end{tabular}%
}
\vspace{-2em}
\end{table}

%% file: sec/5_conclusion.tex
\section{Conclusion}
In this paper, we propose \textbf{\method}, a unified approach for metric depth prediction and pose estimation through spatial-temporal-semantic information fusion for surrounding cameras. By designing a unified transformer architecture, we effectively integrate these features, improving computational efficiency and reducing boundary ambiguity. We also introduce a joint pose estimation network for surrounding cameras that combines depth predictions with camera extrinsic parameters to achieve accurate scale estimation. Additionally, we propose a curvature loss function, where guidance from the depth estimation world model significantly accelerates convergence and improves depth prediction accuracy. Experimental results on two widely used datasets demonstrate the superiority of our method, showcasing its broad applicability in autonomous driving systems with surrounding cameras.